\pdfoutput=1

\documentclass[11pt]{article}

\usepackage[preprint]{acl}

\usepackage{times}
\usepackage{latexsym}

\usepackage[T1]{fontenc}

\usepackage[utf8]{inputenc}

\usepackage{microtype}

\usepackage{inconsolata}

\usepackage{graphicx}

\title{Agent-based Automated Claim Matching with Instruction-following LLMs}

\author{Dina Pisarevskaya \and Arkaitz Zubiaga \\
         Queen Mary University of London, UK \\ \{d.pisarevskaya, a.zubiaga\}@qmul.ac.uk}

\begin{document}
\maketitle
\begin{abstract}
We present a novel agent-based approach for the automated claim matching task with instruction-following LLMs. 
We propose a two-step pipeline that first generates prompts with LLMs, to then perform claim matching as a binary classification task with LLMs. We demonstrate that LLM-generated prompts can outperform SOTA with human-generated prompts, and that smaller LLMs can do as well as larger ones in the generation process, allowing to save computational resources. We also demonstrate the effectiveness of using different LLMs for each step of the pipeline, i.e. using an LLM for prompt generation, and another for claim matching. Our investigation into the prompt generation process in turn reveals insights into the LLMs' understanding of claim matching. 
\end{abstract}

\section{Introduction}
As part of an automated fact-checking pipeline, the claim matching (CM) component determines if two claims (factual statements) can be verified using the same piece of evidence or fact-check~\citep{ceddeno2020, shaar-etal-2020-known, zeng2021automated, guo-etal-2022-survey, pikuliak-etal-2023-multilingual, PANCHENDRARAJAN2024100066}. \citet{pisarevskaya-zubiaga-2025-zero} introduced CM as a binary classification task, proposing state-of-the-art (SOTA) few-shot learning results with four large language models (LLMs), based on hand-crafted prompt templates and few-shot examples, without automated prompt engineering. We aim to further investigate and develop the proposed instruction-following LLMs application for CM, overcoming this limitation. We propose, for the first time, a novel agent-based approach specifically for CM, which outperforms these SOTA results for CM as well as results after SOTA prompt tuning. 

Our contributions are: (1) we investigate automated prompt engineering methods for CM, searching for best few-shot examples and prompt templates; (2) to automate the CM task, we propose an LLM agent-based pipeline to generate specific CM prompts and then use them for the task, comparing performance against SOTA CM approach and the SOTA prompt tuning method; (3) we study the LLMs' understanding of CM, revealing which prompts work better and why. 

\section{Related Work and Motivation}

\textbf{Claim matching} is useful for identifying claims that can be fact-checked together. 
It is addressed as a ranking task~\citep{shaar-etal-2020-known, kazemi-etal-2021-claim, kazemi2022, shaar2022} or, more recently, as a classification task (3 classes recognising textual entailment)~\citep{10.1145/3589335.3651910, 10.1145/3589335.3651504}. \citet{pisarevskaya-zubiaga-2025-zero} investigated manually generated prompt templates with a single user instruction by framing the task as a paraphrase detection, claim matching or natural language inference task, to assess the applicability of closely related tasks to CM.

\textbf{LLM agents} are able to interact, complete tasks, draw conclusions~\citep{10.1609/aaai.v38i17.29936, wangllmagents, li-2025-review}. LLM multi-agent systems use capabilities of single agents, enhancing their collaboration on complex tasks~\citep{ijcai2024p890, han2025llmmultiagentsystemschallenges, liang-etal-2024-encouraging}. 
Inspired by~\citet{Chan2023ChatEvalTB} and \citet{fang-etal-2025-collaborative}, we apply a pipeline interaction between two agents: the output of the first agent becomes the next agent's input. But, for CM, we suggest that the final label depends only on the second agent. 
\textbf{Automated prompt engineering} allows optimising prompts by reducing human effort 
~\citep{Liu2021PretrainPA, schulhoff2025promptreportsystematicsurvey}. We investigate how an automated choice of few-shot examples for a CM prompt can lead to improved performance. \textbf{Generating prompts with LLMs} is an emergent approach~\citep{10.1145/3411763.3451760,zhou2023large, ye-etal-2024-prompt}. \textbf{Prompt-tuning} is an automated method to fine-tune only a small number of model parameters for a task~\citep{lester-etal-2021-power, liu-etal-2022-p, xiao-etal-2023-decomposed, yang-etal-2025-position}. This SOTA is compared to our approach that includes automated prompt generation and few-shot learning. 

\section{Dataset and Experimental Setup}

We use the ClaimMatch dataset~\citep{pisarevskaya-zubiaga-2025-zero}, which is based on~\citet{nakov-clef-2022}. Its test set comprises 500 matching and 500 not matching claim pairs. For consistency, we started our experiments using the few-shot train examples listed in~\citet{pisarevskaya-zubiaga-2025-zero}. For prompt tuning, the remaining 1,682 claim pairs were taken (with train \& validation split 0.8 \& 0.2). 

We chose two open-access instruction-following LLMs for our experiments: Mistral and Llama. To compare the proposed approach with the previous SOTA in the same settings and to ensure reproducibility, we implemented the same model versions: Mistral-7B-Instruct-v0.3 
(Mistral) and Llama-3-8B-Instruct 
(Llama) (max. new tokens 400). The same models were used for prompt tuning (5 epochs, AdamW optimiser, lr 3e-2) with PEFT~\citep{peft}. For prompt generation, we use two bigger models: Mistral-Small-24B-Instruct-2501 
(Mistral-b) and Llama-3.3-70B-Instruct 
(Llama-b), both downloaded with 4-bit quantization to fit the GPU memory 40GB. 

\section{Methodology}

First, we describe experiments on automated choice of few-shot examples for a prompt, to define which train examples to use. After that, we propose the pipeline of LLM agent-based few-shot CM that consists of two parts: automated prompt generation, and claim pairs classification if they match. 

\textbf{Choosing few-shot examples for a prompt}. The three best hand-crafted prompt templates from~\citet{pisarevskaya-zubiaga-2025-zero} were taken as a basis: CM-t, PD-t, and NLI-t (see Table \ref{tab:hc_templates} in Appendix~\ref{sec: appendix}). We kept the focus on structured prompts with 10 balanced few-shot examples. We examined three options of choosing few-shot instances for a prompt and compared them to the previous SOTA results with their manual choice: (1) random choice: 10 random claim pairs (5 positive and 5 negative class examples); (2) choice based on highest/lowest semantic similarity scores: developing the negative examples selection method for the train set in~\citet{kazemi2022}, 5 claim pairs with highest semantic similarity scores were selected as positives, and 5 claim pairs with lowest semantic similarity scores were chosen as negatives; (3) choice with borderline semantic similarity scores added: for both positives and negatives, we took 2 examples with the highest and with the lowest scores for their class, and 1 example with the average semantic similarity score for their class, finally obtaining 5 positives and 5 negatives. We used the All-MiniLM-L6-v2 model for semantic similarity.

\textbf{Generating  prompts with LLMs}. As an initial step, a user prompt with few-shot examples is given to an LLM. It explicitly mentions that the examples contain statement pairs that match or not, without any further details as to how the claim matching task should be tackled. The specific system prompt was added too: ``You are a large language model. Create the best prompt for you for the described task." Provided with the same 10 few-shot examples used in other experiments, the models were aimed to generate a new prompt for a large language model for the task described in the examples. Prompts created with Mistral, if almost similar in their wording, were merged into one prompt. Prompts generated with Llama were also not diverse: the model suggests a definition for the CM task –- whether two statements describe the same event / topic / situation / idea / issue / concept. After an extensive series of experiments, we chose 5 prompts generated with each model. We also generated prompts with bigger models Llama-b and Mistral-b, to make use of a more capable LLM to generate a prompt that will then be used by a smaller LLM, to save resources. The resulting prompts are shown in Table \ref{tab:templates_long} in Appendix~\ref{sec: appendix}. 

\textbf{Claim matching with LLMs}. To assess if the proposed prompts are suitable for CM and can be generalised to other models, we ran few-shot experiments with the prompts created with the models (with 10 few-shot examples added), handling CM as a binary classification task and using four setups: (1) Mistral with Mistral prompts; (2) Mistral with Llama prompts; (3) Llama with Llama prompts; (4) Llama with Mistral prompts. We also investigated if bigger LLMs are better agents to generate a prompt for a smaller LLM from the same model family: if prompts generated with a bigger Mistral-b model are suitable for a smaller Mistral model, and if prompts generated with a bigger Llama-b model are suitable for a smaller Llama model. Results are compared to the results with the overall best CM-t, PD-t and NLI-t prompt templates (see Appendix \ref{sec: appendix}) from~\citet{pisarevskaya-zubiaga-2025-zero}. To compare the proposed agent-based method for automated prompts generation with the SOTA prompt tuning method, we implemented it for Mistral and Llama models, based on CM-t, PD-t, and NLI-t prompt templates. 

\section{Experiment Results}

\textbf{Few-shot examples selection}. Results in Table \ref{tab:examples} demonstrate that, for CM, there is no common pattern in prompt engineering techniques for different models, demonstrating how train examples for the few shot should be selected. For Mistral, original hand-crafted examples in the few shot still get better results than random, sorted and borderline examples. Sorted examples yield the highest scores with CM-t and PD-t templates for Mistral, but NLI-t template works better with random examples (almost outperforming SOTA for CM). Option with borderline examples does not help this model understand the task. However, for Llama all three approaches outperform the SOTA results, with random examples getting the highest scores for CM-t and PD-t templates (but not for NLI-t template as with Mistral). The choice of few-shot examples is model-specific. Hence, for the next step we continue using hand-crafted few-shot examples. 

\begin{table}[htbp!]
    \centering
    \small
    \begin{tabular}{lcccc}
     \hline
      & \multicolumn{2}{c}{\textbf{Mistral}} & \multicolumn{2}{c}{\textbf{Llama}} \\
     \cline{2-5}
     \textbf{Method} & F1, \% & Acc., \% & F1, \% & Acc., \% \\
     \hline
     \multicolumn{5}{c}{\textbf{CM-t template}} \\
     rand & 85.5 & 85.5 & 83.7 & 84.0 \\
     sort & 86.0 & 86.0 & 79.3 & 80.0 \\
     bord & 84.7 & 84.7 & 78.3 & 79.0 \\
     SOTA & 90.6 & 90.6 & 77.6 & 78.3 \\
     \multicolumn{5}{c}{\textbf{PD-t template}} \\
     rand & 93.1 & 93.1 & 84.0 & 84.0 \\
     sort & 94.1 & 94.1 & 79.4 & 79.8 \\
     bord & 93.1 & 93.1 & 83.3 & 83.3 \\
     SOTA & 95.0 & 95.0 & 60.0 & 64.5 \\
     \multicolumn{5}{c}{\textbf{NLI-t template}} \\
     rand & 88.2 & 88.2 & 67.8 & 70.3 \\
     sort & 86.2 & 86.2 & 86.4 & 86.4 \\
     bord & 85.8 & 85.8 & 81.6 & 81.9 \\
     SOTA & 88.3 & 88.3 & 52.8 & 59.1 \\
    \hline    
    \end{tabular}
    \caption{Selecting few-shot examples for a prompt: random, sorted, borderline and SOTA approaches.}
    \label{tab:examples}
\end{table}

\textbf{Claim matching with LLM-generated prompts}. However, Table \ref{tab:generating_prompts} shows promising results for agent-based prompt generation. Although Mistral continues to demonstrate better scores than Llama, as in the SOTA few-shot experiments, we can highlight: 
for both models, Mistral prompts do not improve over SOTA. But, similarly, both models yield better results using Llama prompts, outperforming their SOTA (especially for Llama). Mistral gets the best score with L4: f1 \& accuracy 96.9 (compared to f1 \& accuracy 95.0 for its SOTA result with PD-t). Llama gets the best score with L2: f1 \& accuracy 94.3, greatly improving its SOTA few-shot results with various prompt templates. For Mistral, L2 is the second-best Llama prompt, but it still outperforms SOTA with f1 \& accuracy 95.3. We can conclude that this prompt - ``Identify whether the two statements are describing the same event, topic, or idea, to determine if the statements match or not match.'' – reveals in the most detailed way the essence of the CM task, which corresponds to the human understanding in the gold labels of the dataset: the same event, topic, or idea in two claims would help detect if they can be verified with the same piece of evidence, and find out if they match. 
Although L1 and L5 prompts are rather similar, results for both Mistral and Llama are better with the L1 prompt. A possible explanation is that the same event, topic, situation, or issue (L1) seems to be more important for CM than the same event, idea, or concept (L5). 
This explains why L2 demonstrates good scores for both models: it incorporates the same event and topic, as well as the request to match claims. 
Longer prompts with more detailed explanations do not get better results, see e.g. L3 and M5 results. 
Prompts proposed with different LLMs contain similar core, essential statements reflecting the LLMs understanding of CM: matching claims refer to the same event. It is close to PD-t template, explaining its high performance. 

\begin{table*}[htb]
    \centering
    \small
    \begin{tabular}{lccclcclccclcc}
    \hline
     \multicolumn{10}{c}{\textbf{Prompts generated with Llama models}} & & \multicolumn{3}{c}{\textbf{SOTA approaches}}\\
     \hline
     \multicolumn{6}{c}{\textbf{Llama-generated}} & & \multicolumn{3}{c}{\textbf{Llama-b-generated}} & & \multicolumn{3}{c}{\textbf{Hand-crafted prompts}} \\
     \cline{1-6}
     \cline{8-10}
     \cline{12-14}
      & \multicolumn{2}{c}{\textbf{Llama}} & & \multicolumn{2}{c}{\textbf{Mistral}} & & \multicolumn{3}{c}{\textbf{Llama}} & & Setup & F1, \% & Acc., \% \\
     \cline{12-14}    
     Prompt & F1, \% & Acc., \% & & F1, \% & Acc., \% & & Prompt & F1, \% & Acc., \% & & L CM-t & 77.6 & 78.3 \\
     \cline{1-3}
     \cline{5-6}
     \cline{8-10}
     L1 & 89.9 & 90.0 & & 95.0 & 95.0 & & Lb1 & 59.7 & 64.7 & & L PD-t & 60.0 & 64.5 \\
     L2 & \textbf{94.3} & \textbf{94.3} & & 95.3 & 95.3 & & Lb2 & 62.6 & 64.9 & & L NLI-t & 52.8 & 59.1 \\
     L3 & 79.3 & 79.4 & & 94.9 & 94.9 & & Lb3 & 79.2 & 79.5 & & M CM-t & 90.6 & 90.6 \\     
     L4 & 76.6 & 77.0 & & \textbf{96.9} & \textbf{96.9} & & Lb4 & \textbf{88.6} & \textbf{88.7} & & M PD-t & \textbf{95.0} & \textbf{95.0} \\
     L5 & 88.4 & 88.4 & & 94.6 & 94.6 & & Lb5 & 88.4 & 88.4 & & M NLI-t & 88.3 & 88.3 \\
     \hline
     \multicolumn{10}{c}{\textbf{Prompts generated with Mistral models}} & & & & \\ 
     \hline
     \multicolumn{6}{c}{\textbf{Mistral-generated}} & & \multicolumn{3}{c}{\textbf{Mistral-b-generated}} & & \multicolumn{3}{c}{\textbf{Prompt tuning}} \\
     \cline{1-6}
     \cline{8-10}   
     \cline{12-14}
      & \multicolumn{2}{c}{\textbf{Llama}} & & \multicolumn{2}{c}{\textbf{Mistral}} & & \multicolumn{3}{c}{\textbf{Mistral}} & & Setup & F1, \% & Acc., \%\\     
     \cline{12-14}
     Prompt & F1, \% & Acc., \% & & F1, \% & Acc., \% & & Prompt & F1, \% & Acc., \% & & L CM-t & 57.4 & 59.0 \\
     \cline{1-3}
     \cline{5-6}
     \cline{8-10}     
     M1 & 76.8 & 77.5 & & 82.9 & 83.3 & & Mb1 & 95.1 & 95.1 & & L PD-t & 77.3 & 77.4 \\
     M2 & 54.3 & 61.0 & & \textbf{92.5} & \textbf{92.5} & & Mb2 & 92.6 & 92.6 & & L NLI-t & 59.0 & 62.5 \\
     M3 & 68.7 & 69.1 & & 87.9 & 88.0 & & Mb3 & 91.0 & 91.0 & & M CM-t & 89.0 & 89.1 \\
     M4 & \textbf{77.6} & \textbf{77.6} & & 89.9 & 89.9 & & Mb4 & \textbf{96.2} & \textbf{96.2} & & M PD-t & \textbf{96.4} & \textbf{96.4} \\
     M5 & 61.7 & 63.4 & & 85.8 & 85.8 & & Mb5 & 93.0 & 93.0 & & M NLI-t & 82.2 & 82.7 \\
     \hline
    \end{tabular}
    \caption{Few-shot performance with generated prompts (left and center). SOTA performance (right).}
    \label{tab:generating_prompts}
\end{table*}

Bigger models do not understand the task better than the smaller models: results with their generated prompts are not better than with prompts from a smaller model (Llama). As for Mistral, only two prompt templates (Mb1 and Mb4) yield better results than SOTA for CM. Prompts generated with a bigger Llama-b model do not essentially differ much in their template pattern and core requirements from the prompts from a smaller Llama model. But all prompts from a bigger Mistral-b model have a more specific structured template pattern compared to Mistral, with clear guidelines about more details, placed after the core requirements about the same or similar information in two statements. 
It improved the performance specifically for Mistral model (from the same models family), but still did not outperform Mistral results with prompts from the smaller Llama. A smaller model can be used as an agent to generate a claim matching prompt, saving computational resources. Prompts generated with one LLM can work well for another LLM: Llama is better in creating prompts, then they should be passed to Mistral which is better in CM classification. 
Our approach also outperforms the SOTA prompt tuning approach with the best hand-crafted prompt templates (Mistral with L4 performs better than with PD-t template after prompt tuning). 
Using one LLM to generate CM prompts for another LLM outperforms SOTA results for the task and saves time and resources.

\textbf{Error Analysis.} \emph{The ``same event" requirement in LLMs prompts is not fully reliable}. Both Mistral and Llama yield worse results with Mistral prompts. As class labels are mostly followed with explanations, we can conclude that there are two main types of errors: causing false positives (trying to find a logical connection between unrelated claims in the negative class examples) and false negatives (if two statements in a claim pair vary in some non-substantial, or significant details, a model creates logical connections, leading to wrong conclusions). 
Prompt templates M2 and M3 can make a model yield a negative class label, if two claims vary in some non-substantial details. Hence, ``the same event'' is not a fully reliable and sufficient marker to detect if two claims in a pair match. Llama prompts, generally, contain more additional requirements than Mistral prompts: it is not only the same event, but can also be the same topic, situation, issue, idea, or concept. There are examples where such requirements are too strict: ``a similar, but not the exact same  event''. 
But such broad requirements can still help reduce the number of false negatives: ``the same but from different perspectives''. 

\emph{LLMs prompts assume that contradictory claims cannot match}. In the human understanding of CM, two claims can match if their evidence is the same, even if they contain some contradictory information. Hence, consistency between two claims, proposed as a marker in M3 and, especially, M4 and L3 prompts, does not fully correspond to CM. 
Mistral (with M3) and Llama (with M4) explain their false negative class labels for the same claim pair: ``No, the event or situation described in 1 ([...] granting a wish that `Friends' would stay on Netflix in 2019) is not consistent with the event or situation described in 2 (Netflix announcing that `Friends' would no longer be available [...] after the end of 2018)'' and ``The statements are contradictory. 1 states that `Friends' will still be available on Netflix [...] throughout 2019, while 2 states [...] the show would no longer be available after the end of 2018''. Simple positive class examples with some variations in details are usually classified correctly, but major variations can lead to misclassification.

\section{Conclusion}

We propose a pipeline for agent-based few-shot CM that incorporates a novel approach to automated prompt generation with LLMs. LLM-generated prompts demonstrate that they understand the specifics of the CM task, and with these prompts two LLMs outperform SOTA, based on hand-crafted prompts. We find that a prompt that considers matching claims as those describing the same event, topic or idea performs best, and we show that there is still room for improvement by incorporating additional markers.

\section*{Limitations}

It should be highlighted that we do not aim to compare the results of bigger and smaller models in these experiments. Two bigger models used for prompt generation are not only of larger size, they are also from more recent generations than the smaller models we use, which could explain their better quality. On the other hand, they are applied with quantization, which could reduce their quality. Our aim is to check if prompts, generated with bigger models of the same type, would be useful for smaller models. 

We intentionally provided in the initial prompt, provided to LLMs, that the examples contain statement pairs that
match or do not match, without giving any additional details, or including the definition that matching claims can be verified with the same evidence, or fact-check. Firstly, our aim was to enhance LLMs generated prompts, based on their understanding of the claim matching task (instead of providing human-crafted prompts, that would define claim matching, to them), and evaluate LLMs performance with them. Secondly, claim matching task is closely connected to other tasks in the fact-checking pipeline, but it is a specific task that does not require any fact checking of the claims provided. In the series of experiments, we found out that, if manually created prompts are not carefully curated, LLMs can wrongly address claim matching as the fact-checking task and predict the veracity of two claims instead of the output label if they match / not match. Providing the human definition of claim matching, that incorporates fact-checking purposes, could encourage such model's behaviour. However, as LLMs understanding of the task has its own limitations, further experiments are needed to define better prompts - for example, using step-by-step-reasoning to let a LLM improve prompts in multiple iterations. 
In all Mistral experiments with Llama prompts, recall for the negative class, and precision for the positive class are significantly higher than for another class. An explanation could be that too ``strict'' ``the same event'' requirement in a prompt lead to a significant number of false negatives, where matching claims are marked as not matching, as they vary in some details. On the other hand, a model does not find any non-existent logical connection between two definitely not matching claims (reducing the number of false positives). Step-by-step-reasoning could further improve this issue.

While we kept, for the consistency, the same number of few-shot examples for a prompt, as in the SOTA experiments, further options to automate a choice of few-shot examples for a prompt should be studied, such as their ordering~\citep{lu-etal-2022-fantastically, pmlr-v139-zhao21c}, rephrasing~\citep{yang2024justrephraseituncertainty}, prompt and hyperparameter selection such as the number of labeled examples~\citep{perez2021true}, or choice of different examples for different prompts. 

~\citet{ye-etal-2024-prompt} suggest that a LLM can continuously propose new potentially better prompts. Unlike this, our agentic system exists in the offline mode, to better understand the LLMs understanding of CM as this task can be referred to a few related tasks: the first agent, initialised with a basic prompt and a batch of examples, generates instructions for them, the second agent uses them, then we evaluate the results. Continuos prompt improvement methods, as well as adding classification evaluation with LLMs as the third component of the agent-based CM pipeline, should be investigated in further research. 

We also acknowledge that more research should be done to improve the generated prompts with various prompt engineering techniques (such as chain of thought), as well as test this approach on various datasets and in the multilingual setups. Such LLM-proposed markers of matching claims, as same event and consistency of two claims, can limit the task understanding, and should be further investigated. 

\bibliography{custom}

\begin{thebibliography}{35}
\providecommand{\natexlab}[1]{#1}

\bibitem[{Barrón-Cedeño et~al.(2020)Barrón-Cedeño, Elsayed, Nakov, Martino, Hasanain, Suwaileh, Haouari, Babulkov, Hamdan, Nikolov, Shaar, and Ali}]{ceddeno2020}
Alberto Barrón-Cedeño, Tamer Elsayed, Preslav Nakov, Giovanni Da~San Martino, Maram Hasanain, Reem Suwaileh, Fatima Haouari, Nikolay Babulkov, Bayan Hamdan, Alex Nikolov, Shaden Shaar, and Zien~Sheikh Ali. 2020.
\newblock \href {https://arxiv.org/abs/2007.07997v1} {Overview of checkthat!2020: Automatic identification and verification of claims in social media}.
\newblock \emph{Preprint}, arXiv:2007.07997v1.

\bibitem[{Chan et~al.(2023)Chan, Chen, Su, Yu, Xue, Zhang, Fu, and Liu}]{Chan2023ChatEvalTB}
Chi-Min Chan, Weize Chen, Yusheng Su, Jianxuan Yu, Wei Xue, Shan Zhang, Jie Fu, and Zhiyuan Liu. 2023.
\newblock \href {https://api.semanticscholar.org/CorpusID:260887105} {Chateval: Towards better llm-based evaluators through multi-agent debate}.
\newblock \emph{ArXiv}, abs/2308.07201.

\bibitem[{Choi and Ferrara(2024{\natexlab{a}})}]{10.1145/3589335.3651910}
Eun~Cheol Choi and Emilio Ferrara. 2024{\natexlab{a}}.
\newblock \href {https://doi.org/10.1145/3589335.3651910} {Automated claim matching with large language models: Empowering fact-checkers in the fight against misinformation}.
\newblock In \emph{Companion Proceedings of the ACM Web Conference 2024}, WWW '24, page 1441–1449, New York, NY, USA. Association for Computing Machinery.

\bibitem[{Choi and Ferrara(2024{\natexlab{b}})}]{10.1145/3589335.3651504}
Eun~Cheol Choi and Emilio Ferrara. 2024{\natexlab{b}}.
\newblock \href {https://doi.org/10.1145/3589335.3651504} {Fact-gpt: Fact-checking augmentation via claim matching with llms}.
\newblock In \emph{Companion Proceedings of the ACM Web Conference 2024}, WWW '24, page 883–886, New York, NY, USA. Association for Computing Machinery.

\bibitem[{Fang et~al.(2025)Fang, Qiang, Ouyang, Zhu, Yuan, and Li}]{fang-etal-2025-collaborative}
Dengzhao Fang, Jipeng Qiang, Xiaoye Ouyang, Yi~Zhu, Yunhao Yuan, and Yun Li. 2025.
\newblock \href {https://aclanthology.org/2025.coling-main.60/} {Collaborative document simplification using multi-agent systems}.
\newblock In \emph{Proceedings of the 31st International Conference on Computational Linguistics}, pages 897--912, Abu Dhabi, UAE. Association for Computational Linguistics.

\bibitem[{Guo et~al.(2024)Guo, Chen, Wang, Chang, Pei, Chawla, Wiest, and Zhang}]{ijcai2024p890}
Taicheng Guo, Xiuying Chen, Yaqi Wang, Ruidi Chang, Shichao Pei, Nitesh~V. Chawla, Olaf Wiest, and Xiangliang Zhang. 2024.
\newblock \href {https://doi.org/10.24963/ijcai.2024/890} {Large language model based multi-agents: A survey of progress and challenges}.
\newblock In \emph{Proceedings of the Thirty-Third International Joint Conference on Artificial Intelligence, {IJCAI-24}}, pages 8048--8057. International Joint Conferences on Artificial Intelligence Organization.
\newblock Survey Track.

\bibitem[{Guo et~al.(2022)Guo, Schlichtkrull, and Vlachos}]{guo-etal-2022-survey}
Zhijiang Guo, Michael Schlichtkrull, and Andreas Vlachos. 2022.
\newblock \href {https://doi.org/10.1162/tacl_a_00454} {A survey on automated fact-checking}.
\newblock \emph{Transactions of the Association for Computational Linguistics}, 10:178--206.

\bibitem[{Han et~al.(2025)Han, Zhang, Yao, Jin, and Xu}]{han2025llmmultiagentsystemschallenges}
Shanshan Han, Qifan Zhang, Yuhang Yao, Weizhao Jin, and Zhaozhuo Xu. 2025.
\newblock \href {https://arxiv.org/abs/2402.03578} {Llm multi-agent systems: Challenges and open problems}.
\newblock \emph{Preprint}, arXiv:2402.03578.

\bibitem[{Kazemi et~al.(2021)Kazemi, Garimella, Gaffney, and Hale}]{kazemi-etal-2021-claim}
Ashkan Kazemi, Kiran Garimella, Devin Gaffney, and Scott~A. Hale. 2021.
\newblock \href {https://doi.org/10.18653/v1/2021.acl-long.347} {Claim matching beyond {E}nglish to scale global fact-checking}.
\newblock In \emph{Proceedings of the 59th Annual Meeting of the Association for Computational Linguistics and the 11th International Joint Conference on Natural Language Processing (Volume 1: Long Papers)}, pages 4504--4517, Online. Association for Computational Linguistics.

\bibitem[{Kazemi et~al.(2022)Kazemi, Li, Perez-Rosas, Hale, and Mihalcea}]{kazemi2022}
Ashkan Kazemi, Zehua Li, Veronica Perez-Rosas, Scott~A. Hale, and Rada Mihalcea. 2022.
\newblock \href {https://arxiv.org/abs/2202.07094} {Matching tweets with applicable fact-checks across languages}.
\newblock \emph{Preprint}, arXiv:2202.07094.

\bibitem[{Lester et~al.(2021)Lester, Al-Rfou, and Constant}]{lester-etal-2021-power}
Brian Lester, Rami Al-Rfou, and Noah Constant. 2021.
\newblock \href {https://doi.org/10.18653/v1/2021.emnlp-main.243} {The power of scale for parameter-efficient prompt tuning}.
\newblock In \emph{Proceedings of the 2021 Conference on Empirical Methods in Natural Language Processing}, pages 3045--3059, Online and Punta Cana, Dominican Republic. Association for Computational Linguistics.

\bibitem[{Li(2025)}]{li-2025-review}
Xinzhe Li. 2025.
\newblock \href {https://aclanthology.org/2025.coling-main.652/} {A review of prominent paradigms for {LLM}-based agents: Tool use, planning (including {RAG}), and feedback learning}.
\newblock In \emph{Proceedings of the 31st International Conference on Computational Linguistics}, pages 9760--9779, Abu Dhabi, UAE. Association for Computational Linguistics.

\bibitem[{Liang et~al.(2024)Liang, He, Jiao, Wang, Wang, Wang, Yang, Shi, and Tu}]{liang-etal-2024-encouraging}
Tian Liang, Zhiwei He, Wenxiang Jiao, Xing Wang, Yan Wang, Rui Wang, Yujiu Yang, Shuming Shi, and Zhaopeng Tu. 2024.
\newblock \href {https://doi.org/10.18653/v1/2024.emnlp-main.992} {Encouraging divergent thinking in large language models through multi-agent debate}.
\newblock In \emph{Proceedings of the 2024 Conference on Empirical Methods in Natural Language Processing}, pages 17889--17904, Miami, Florida, USA. Association for Computational Linguistics.

\bibitem[{Liu et~al.(2021)Liu, Yuan, Fu, Jiang, Hayashi, and Neubig}]{Liu2021PretrainPA}
Pengfei Liu, Weizhe Yuan, Jinlan Fu, Zhengbao Jiang, Hiroaki Hayashi, and Graham Neubig. 2021.
\newblock \href {https://api.semanticscholar.org/CorpusID:236493269} {Pre-train, prompt, and predict: A systematic survey of prompting methods in natural language processing}.
\newblock \emph{ACM Computing Surveys}, 55:1 -- 35.

\bibitem[{Liu et~al.(2022)Liu, Ji, Fu, Tam, Du, Yang, and Tang}]{liu-etal-2022-p}
Xiao Liu, Kaixuan Ji, Yicheng Fu, Weng Tam, Zhengxiao Du, Zhilin Yang, and Jie Tang. 2022.
\newblock \href {https://doi.org/10.18653/v1/2022.acl-short.8} {{P}-tuning: Prompt tuning can be comparable to fine-tuning across scales and tasks}.
\newblock In \emph{Proceedings of the 60th Annual Meeting of the Association for Computational Linguistics (Volume 2: Short Papers)}, pages 61--68, Dublin, Ireland. Association for Computational Linguistics.

\bibitem[{Lu et~al.(2022)Lu, Bartolo, Moore, Riedel, and Stenetorp}]{lu-etal-2022-fantastically}
Yao Lu, Max Bartolo, Alastair Moore, Sebastian Riedel, and Pontus Stenetorp. 2022.
\newblock \href {https://doi.org/10.18653/v1/2022.acl-long.556} {Fantastically ordered prompts and where to find them: Overcoming few-shot prompt order sensitivity}.
\newblock In \emph{Proceedings of the 60th Annual Meeting of the Association for Computational Linguistics (Volume 1: Long Papers)}, pages 8086--8098, Dublin, Ireland. Association for Computational Linguistics.

\bibitem[{Mangrulkar et~al.(2022)Mangrulkar, Gugger, Debut, Belkada, Paul, and Bossan}]{peft}
Sourab Mangrulkar, Sylvain Gugger, Lysandre Debut, Younes Belkada, Sayak Paul, and Benjamin Bossan. 2022.
\newblock Peft: State-of-the-art parameter-efficient fine-tuning methods.
\newblock \url{https://github.com/huggingface/peft}.

\bibitem[{Nakov et~al.(2022)Nakov, Martino, Alam, Shaar, Mubarak, and Babulkov}]{nakov-clef-2022}
Preslav Nakov, Giovanni Da~San Martino, Firoj Alam, Shaden Shaar, Hamdy Mubarak, and Nikolay Babulkov. 2022.
\newblock Overview of the {CLEF}-2022 {C}heck{T}hat! lab task 2 on detecting previously fact-checked claims.
\newblock In \emph{Proceedings of the Working Notes of CLEF 2022 - Conference and Labs of the Evaluation Forum}.

\bibitem[{Panchendrarajan and Zubiaga(2024)}]{PANCHENDRARAJAN2024100066}
Rrubaa Panchendrarajan and Arkaitz Zubiaga. 2024.
\newblock \href {https://doi.org/10.1016/j.nlp.2024.100066} {Claim detection for automated fact-checking: A survey on monolingual, multilingual and cross-lingual research}.
\newblock \emph{Natural Language Processing Journal}, 7:100066.

\bibitem[{Perez et~al.(2021)Perez, Kiela, and Cho}]{perez2021true}
Ethan Perez, Douwe Kiela, and Kyunghyun Cho. 2021.
\newblock \href {https://arxiv.org/abs/2105.11447} {True few-shot learning with language models}.
\newblock \emph{NeurIPS}.

\bibitem[{Pikuliak et~al.(2023)Pikuliak, Srba, Moro, Hromadka, Smole{\v{n}}, Meli{\v{s}}ek, Vykopal, Simko, Podrou{\v{z}}ek, and Bielikova}]{pikuliak-etal-2023-multilingual}
Mat{\'u}{\v{s}} Pikuliak, Ivan Srba, Robert Moro, Timo Hromadka, Timotej Smole{\v{n}}, Martin Meli{\v{s}}ek, Ivan Vykopal, Jakub Simko, Juraj Podrou{\v{z}}ek, and Maria Bielikova. 2023.
\newblock \href {https://doi.org/10.18653/v1/2023.emnlp-main.1027} {Multilingual previously fact-checked claim retrieval}.
\newblock In \emph{Proceedings of the 2023 Conference on Empirical Methods in Natural Language Processing}, pages 16477--16500, Singapore. Association for Computational Linguistics.

\bibitem[{Pisarevskaya and Zubiaga(2025)}]{pisarevskaya-zubiaga-2025-zero}
Dina Pisarevskaya and Arkaitz Zubiaga. 2025.
\newblock \href {https://aclanthology.org/2025.coling-main.650/} {Zero-shot and few-shot learning with instruction-following {LLM}s for claim matching in automated fact-checking}.
\newblock In \emph{Proceedings of the 31st International Conference on Computational Linguistics}, pages 9721--9736, Abu Dhabi, UAE. Association for Computational Linguistics.

\bibitem[{Reynolds and McDonell(2021)}]{10.1145/3411763.3451760}
Laria Reynolds and Kyle McDonell. 2021.
\newblock \href {https://doi.org/10.1145/3411763.3451760} {Prompt programming for large language models: Beyond the few-shot paradigm}.
\newblock In \emph{Extended Abstracts of the 2021 CHI Conference on Human Factors in Computing Systems}, CHI EA '21, New York, NY, USA. Association for Computing Machinery.

\bibitem[{Schulhoff et~al.(2025)Schulhoff, Ilie, Balepur, Kahadze, Liu, Si, Li, Gupta, Han, Schulhoff, Dulepet, Vidyadhara, Ki, Agrawal, Pham, Kroiz, Li, Tao, Srivastava, Costa, Gupta, Rogers, Goncearenco, Sarli, Galynker, Peskoff, Carpuat, White, Anadkat, Hoyle, and Resnik}]{schulhoff2025promptreportsystematicsurvey}
Sander Schulhoff, Michael Ilie, Nishant Balepur, Konstantine Kahadze, Amanda Liu, Chenglei Si, Yinheng Li, Aayush Gupta, HyoJung Han, Sevien Schulhoff, Pranav~Sandeep Dulepet, Saurav Vidyadhara, Dayeon Ki, Sweta Agrawal, Chau Pham, Gerson Kroiz, Feileen Li, Hudson Tao, Ashay Srivastava, and 12 others. 2025.
\newblock \href {https://arxiv.org/abs/2406.06608} {The prompt report: A systematic survey of prompt engineering techniques}.
\newblock \emph{Preprint}, arXiv:2406.06608.

\bibitem[{Shaar et~al.(2022)Shaar, Alam, Da~San~Martino, and Nakov}]{shaar2022}
Shaden Shaar, Firoj Alam, Giovanni Da~San~Martino, and Preslav Nakov. 2022.
\newblock \href {https://doi.org/10.18653/v1/2022.findings-naacl.122} {The role of context in detecting previously fact-checked claims}.
\newblock In \emph{Findings of the Association for Computational Linguistics: NAACL 2022}, pages 1619--1631, Seattle, United States. Association for Computational Linguistics.

\bibitem[{Shaar et~al.(2020)Shaar, Babulkov, Da~San~Martino, and Nakov}]{shaar-etal-2020-known}
Shaden Shaar, Nikolay Babulkov, Giovanni Da~San~Martino, and Preslav Nakov. 2020.
\newblock \href {https://doi.org/10.18653/v1/2020.acl-main.332} {That is a known lie: Detecting previously fact-checked claims}.
\newblock In \emph{Proceedings of the 58th Annual Meeting of the Association for Computational Linguistics}, pages 3607--3618, Online. Association for Computational Linguistics.

\bibitem[{Wang et~al.(2024)Wang, Ma, Feng, Zhang, Yang, Zhang, Chen, Tang, Chen, Lin, Zhao, Wei, and Wen}]{wangllmagents}
Lei Wang, Chen Ma, Xueyang Feng, Zeyu Zhang, Hao Yang, Jingsen Zhang, Zhiyuan Chen, Jiakai Tang, Xu~Chen, Yankai Lin, Wayne~Xin Zhao, Zhewei Wei, and Jirong Wen. 2024.
\newblock \href {https://doi.org/10.1007/s11704-024-40231-1} {A survey on large language model based autonomous agents}.
\newblock \emph{Frontiers of Computer Science}, 18.

\bibitem[{Xiao et~al.(2023)Xiao, Xu, Li, Lu, and Li}]{xiao-etal-2023-decomposed}
Yao Xiao, Lu~Xu, Jiaxi Li, Wei Lu, and Xiaoli Li. 2023.
\newblock \href {https://doi.org/10.18653/v1/2023.findings-emnlp.890} {Decomposed prompt tuning via low-rank reparameterization}.
\newblock In \emph{Findings of the Association for Computational Linguistics: EMNLP 2023}, pages 13335--13347, Singapore. Association for Computational Linguistics.

\bibitem[{Yang et~al.(2024)Yang, Chen, and Pitas}]{yang2024justrephraseituncertainty}
Adam Yang, Chen Chen, and Konstantinos Pitas. 2024.
\newblock \href {https://arxiv.org/abs/2405.13907} {Just rephrase it! uncertainty estimation in closed-source language models via multiple rephrased queries}.
\newblock \emph{Preprint}, arXiv:2405.13907.

\bibitem[{Yang et~al.(2025)Yang, Cheng, Zhao, Yu, Petzold, and Chen}]{yang-etal-2025-position}
Xianjun Yang, Wei Cheng, Xujiang Zhao, Wenchao Yu, Linda~Ruth Petzold, and Haifeng Chen. 2025.
\newblock \href {https://aclanthology.org/2025.findings-naacl.474/} {Position really matters: Towards a holistic approach for prompt tuning}.
\newblock In \emph{Findings of the Association for Computational Linguistics: NAACL 2025}, pages 8501--8523, Albuquerque, New Mexico. Association for Computational Linguistics.

\bibitem[{Ye et~al.(2024)Ye, Ahmed, Pryzant, and Khani}]{ye-etal-2024-prompt}
Qinyuan Ye, Mohamed Ahmed, Reid Pryzant, and Fereshte Khani. 2024.
\newblock \href {https://doi.org/10.18653/v1/2024.findings-acl.21} {Prompt engineering a prompt engineer}.
\newblock In \emph{Findings of the Association for Computational Linguistics: ACL 2024}, pages 355--385, Bangkok, Thailand. Association for Computational Linguistics.

\bibitem[{Zeng et~al.(2021)Zeng, Abumansour, and Zubiaga}]{zeng2021automated}
Xia Zeng, Amani~S Abumansour, and Arkaitz Zubiaga. 2021.
\newblock Automated fact-checking: A survey.
\newblock \emph{Language and Linguistics Compass}, 15(10):e12438.

\bibitem[{Zhao et~al.(2024)Zhao, Huang, Xu, Lin, Liu, and Huang}]{10.1609/aaai.v38i17.29936}
Andrew Zhao, Daniel Huang, Quentin Xu, Matthieu Lin, Yong-Jin Liu, and Gao Huang. 2024.
\newblock \href {https://doi.org/10.1609/aaai.v38i17.29936} {Expel: Llm agents are experiential learners}.
\newblock In \emph{Proceedings of the Thirty-Eighth AAAI Conference on Artificial Intelligence and Thirty-Sixth Conference on Innovative Applications of Artificial Intelligence and Fourteenth Symposium on Educational Advances in Artificial Intelligence}, AAAI'24/IAAI'24/EAAI'24. AAAI Press.

\bibitem[{Zhao et~al.(2021)Zhao, Wallace, Feng, Klein, and Singh}]{pmlr-v139-zhao21c}
Zihao Zhao, Eric Wallace, Shi Feng, Dan Klein, and Sameer Singh. 2021.
\newblock \href {https://proceedings.mlr.press/v139/zhao21c.html} {Calibrate before use: Improving few-shot performance of language models}.
\newblock In \emph{Proceedings of the 38th International Conference on Machine Learning}, volume 139 of \emph{Proceedings of Machine Learning Research}, pages 12697--12706. PMLR.

\bibitem[{Zhou et~al.(2023)Zhou, Muresanu, Han, Paster, Pitis, Chan, and Ba}]{zhou2023large}
Yongchao Zhou, Andrei~Ioan Muresanu, Ziwen Han, Keiran Paster, Silviu Pitis, Harris Chan, and Jimmy Ba. 2023.
\newblock \href {https://openreview.net/forum?id=92gvk82DE-} {Large language models are human-level prompt engineers}.
\newblock In \emph{The Eleventh International Conference on Learning Representations}.

\end{thebibliography}

\appendix

\section{Appendix}

\label{sec: appendix}

\begin{table}[htbp!]
    \centering
    \small
    \begin{tabular}{l}
     \hline
     \textbf{CM-t template} \\
     \hline
     A Matches to B. 
     Correct? Answer: [yes/no] \\
     \hline
     \textbf{PD-t template} \\
     \hline
     A. B. Question: Do A and B both refer to \\
     the same event? Yes or no? Answer: [yes/no] \\
     \hline
     \textbf{NLI-t template} \\
     \hline
     Suppose A. Can we infer that B? \\
     Yes or no? Answer: [yes/no] \\
     \hline
    \end{tabular}
    \caption{Hand-crafted prompt templates.}
    \label{tab:hc_templates}
\end{table}

The best performing hand-crafted claim matching (CM-t), paraphrase detection (PD-t) and natural language understanding (NLI-t) prompt templates from~\citet{pisarevskaya-zubiaga-2025-zero}, used in the experiments, are provided in Table \ref{tab:hc_templates}.

The full prompts generated with the models are presented in Table \ref{tab:templates_long}. The prompts were generated with smaller models (Mistral, Llama) and bigger models (Mistral-b, Llama-b). To use these prompts for claim matching classification with LLMs, they were processed for each LLM to adhere to its specific requirements to a prompt template. 10 few-shot examples were added to each prompt. 

\begin{table*}[htb]
    \centering
    \small
    \begin{tabular}{p{1\linewidth}}   
     \hline
     \textbf{Prompts generated with Mistral} \\
     \hline
      1. In the next example, do statements 1 and 2 both discuss or mention the same event or phenomenon? \\
      2. Determine if Statement 1 and Statement 2 describe the same event or not. If they describe the same event, the answer should be "yes," otherwise it should be "no." \\
      3. Is the event or situation described in Statement 1 consistent with the event or situation described in Statement 2? If yes, then the correct answer should be "yes." If not, then the correct answer should be "no." \\   
      4. Given two statements, determine if the information provided in both statements is consistent or contradictory. State whether the statements Match or Do Not Match. \\
      5. Given two statements, determine if they are logically consistent or contradictory. The answer should be in the format of "match" or "not match". For example, if two statements are about the same event, person, or object, and they agree, the answer would be "match". If they disagree, the answer would be "not match". The statements should be factual and verifiable, and no assumptions should be made about the intention or interpretation of the statements. \\
      \hline
      \textbf{Prompts generated with Llama} \\
      \hline
      1. Do the two statements describe the same event, topic, situation, or issue? \\
      2. Identify whether the two statements are describing the same event, topic, or idea, to determine if the statements match or not match. \\
      3. Given two statements,  determine whether they are describing the same event, idea, or concept, or if they are unrelated. Please provide a binary answer: 'yes' if the statements match, or 'no' if they do not match. The statements may be from different sources, and the model should rely on its understanding of the content to make a decision. Examples of matching statements include identical events or ideas, while non-matching statements may be unrelated topics, contradictory information, or different perspectives on the same issue. \\
      4. Given two statements, determine if they are describing the same event, topic, or idea. Please indicate whether the statements 'match' or 'do not match' based on the information provided. Use the examples below as a reference for understanding the format and tone of the statements. Please respond with a simple 'yes' or 'no' to indicate whether the statements match or not match. \\
      5. Are the two statements describing the same event, idea, or concept? \\
      \hline
      \textbf{Prompts generated with Mistral-b} \\
      \hline
      1. You will be given two statements and your task is to determine if they match or not. To do this, consider the following guidelines: 1. **Content**: Check if the main subjects, events, or facts described in both statements are the same. 2. **Context**: Consider the context and timeframe of the statements. They should be related to the same event or topic. 3. **Source**: Ignore the source or author of the statements, as they might be paraphrased or reported differently. 4. **Negation**: Be mindful of negations in the statements. A statement and its negation cannot match. 5. **Irrelevant Details**: Ignore minor details or differences in wording that do not affect the core meaning of the statements.  Based on these guidelines, determine if the two statements match or not. Answer with a simple "yes" or "no". \\
      2. You are given two statements and your task is to determine if they match or not. The statements may come from various sources and cover a wide range of topics. Your goal is to assess whether the information conveyed in both statements is consistent and related, or if they are unrelated or contradictory. \#\#\#\# Instructions: 1. Read both statements carefully. 2. Analyze the content and context of each statement. 3. Determine if the statements match in terms of the information they present. 4. Output "yes" if the statements match, and "no" if they do not match. \\
      3. You are given two statements and your task is to determine if they match or not. The statements may come from various sources and cover a wide range of topics. Your goal is to assess whether the information conveyed in both statements is essentially the same or if they are different. \#\#\#\# Instructions: 1. Read both statements carefully. 2. Compare the key information, events, and details mentioned in each statement. 3. Determine if the statements convey the same information or if there are significant differences. 4. Output "yes" if the statements match and "no" if they do not. \\
      4. You are a large language model trained to determine whether two statements match or not match. You will be given two statements and your task is to assess their relevance to each other. The statements may come from various sources and discuss different topics. Your role is to identify if the statements are related, as indicated by whether they refer to the same event, person, or topic, and provide a "yes" if they match and a "no" if they do not. \#\#\# Guidelines: 1. **Focus on Content**: Pay attention to the main subjects, events, and key details mentioned in each statement. 2. **Contextual Relevance**: Determine if the statements are discussing the same context or incident. 3. **Specific Details**: Match specific details such as names, dates, locations, and actions described in the statements. 4. **Logical Consistency**: Ensure that the statements logically align if they are meant to describe the same event or topic. \\
      5. You will be given two statements, labeled Statement 1 and Statement 2. Your task is to determine if the two statements match or not match. To do this, consider the following guidelines: 1. **Matching Statements**: Two statements match if they convey the same or very similar information, even if the phrasing is different. They should agree on the key facts and events described. 2. **Non-Matching Statements**: Two statements do not match if they describe different events, people, or outcomes, or if they present contradictory information.', 'Please provide your answer as either "yes" (for matching statements) or "no" (for non-matching statements). \\ 
      \hline
      \textbf{Prompts generated with Llama-b} \\
      \hline
      1. Compare the information in Statement 1 and Statement 2 to determine if they convey the same information or describe the same event. Consider the context, facts, and details presented in each statement. Output 'yes' if the statements match and 'no' if they do not match. \\
      2. Compare the information presented in Statement 1 and Statement 2. Determine if the main claims, events, or facts described in both statements are consistent with each other. If the statements convey the same overall message, outcome, or conclusion, indicate 'yes'. If the statements contradict each other, present different information, or have distinct conclusions, indicate 'no'. \\
      3. Compare the information presented in Statement 1 and Statement 2. Determine if the two statements convey the same or similar information, or if they contradict each other. Return 'yes' if the statements match and 'no' if they do not match. \\
      4. Analyze the semantic meaning and factual content of two given statements and determine whether they convey the same information, describe the same event, or express the same idea, returning 'yes' if they match and 'no' if they do not match. \\
      5. Do the two statements describe the same event or situation? \\
     \hline
    \end{tabular}
    \caption{Prompts generated for claim matching.}
    \label{tab:templates_long}
\end{table*}

\end{document}